\pdfoutput=1
\documentclass[11pt]{article}
\usepackage[]{ACL2023}
\usepackage{times}
\usepackage{latexsym}
\usepackage{array}
\usepackage{tabularx}
\usepackage{placeins}
\usepackage[T1]{fontenc}
\usepackage[utf8]{inputenc}
\usepackage{microtype}
\usepackage{inconsolata}
\usepackage{float}
\usepackage{scalerel,xparse}
\usepackage{cite}
\usepackage{amsmath,amssymb,amsfonts}
\usepackage{algorithmic}
\usepackage{graphicx}
\usepackage{adjustbox}
\usepackage{textcomp}
\usepackage{multirow}
\usepackage{subfigure}
\usepackage{subcaption}
\usepackage{hyperref}
\usepackage{booktabs}
\usepackage{bbding}
\usepackage{pifont}
\usepackage{xprintlen}
\usepackage[most]{tcolorbox}
\usepackage{xurl}
\usepackage{fancyhdr}

\definecolor{ForestGreen}{RGB}{34, 139, 34}
\definecolor{Salmon}{RGB}{250, 128, 114}
\definecolor{Emerald Green}{RGB}{80, 200, 120}
\definecolor{Canary Yellow}{RGB}{255, 255, 143}
\definecolor{Light Blue}{RGB}{173, 216, 230}
\definecolor{Buff}{RGB}{218, 160, 109}
\definecolor{Light Orange}{RGB}{255, 213, 128}

\newlength{\adjustedwidth}
\fancypagestyle{firststyle}
{
   \fancyhf{}
   \fancyfoot[L]{\textit{Accepted in The 9th Workshop on Noisy and User-generated Text (W-NUT), 18th Conference of the European Chapter of the Association for Computational Linguistics (EACL 2024)}}

}

\title{A Comparative Analysis of Noise Reduction Methods in Sentiment Analysis on Noisy Bangla Texts}

\author{
Kazi Toufique Elahi, Tasnuva Binte Rahman, Shakil Shahriar, Samir Sarker,\\
\textbf{Md. Tanvir Rouf Shawon, G. M. Shahariar}\\
Department of Computer Science and Engineering\\
Ahsanullah University of Science and Technology, Dhaka, Bangladesh\\
\textit{\{ktoufiquee, tasnuvabinterahmansrishti, shakilshahriararnob, rohitsarker5,}\\\textit{shawontanvir95, sshibli745\}@gmail.com}
}

\def\bng{\bngx}

\font\bngx=bang10

\def\*#1*#2{o\null{#2}{#1}}

\def\sh#1{\setbox0=\hbox{#1}%
     \kern-.02em\copy0\kern-\wd0
     \kern.04em\copy0\kern-\wd0
     \kern-.02em\raise.0433em\box0 }
\begin{document}
\maketitle

\thispagestyle{firststyle}
\begin{abstract}
While Bangla is considered a language with limited resources, sentiment analysis has been a subject of extensive research in the literature. Nevertheless, there is a scarcity of exploration into sentiment analysis specifically in the realm of noisy Bangla texts. In this paper, we introduce a dataset (\textbf{NC-SentNoB}) that we annotated manually to identify ten different types of noise found in a pre-existing sentiment analysis dataset comprising of around 15K noisy Bangla texts. At first, given an input noisy text, we identify the noise type, addressing this as a multi-label classification task. Then, we introduce baseline noise reduction methods to alleviate noise prior to conducting sentiment analysis. Finally, we assess the performance of fine-tuned sentiment analysis models with both noisy and noise-reduced texts to make comparisons. The experimental findings indicate that the noise reduction methods utilized are not satisfactory, highlighting the need for more suitable noise reduction methods in future research endeavors. We have made the implementation and dataset presented in this paper publicly available\footnote{\url{https://github.com/ktoufiquee/A-Comparative-Analysis-of-Noise-Reduction-Methods-in-Sentiment-Analysis-on-Noisy-Bangla-Texts}}.
\end{abstract}

\section{Introduction}
Sentiment analysis is the process of analyzing and categorizing the emotions or opinions expressed in textual content. This process holds considerable importance in evaluating public sentiments, analyzing social media posts, and assessing customer feedback. It contributes significantly to gaining insights into ongoing social media dynamics. There have been nearly 7000 papers published on this topic and 99\% of the papers have appeared after 2004, making sentiment analysis one of the fastest-growing research areas \citep{mantyla2018evolution}.\par

\begin{table}[h]
    \centering
    \begin{adjustbox}{width=\columnwidth, center}
    \begin{tabular}{p{0.15\adjustedwidth}p{0.55\adjustedwidth}p{0.3\adjustedwidth}}
        \hline
        \multicolumn{1}{c}{\textbf{Sentiment}} & \multicolumn{1}{c}{\textbf{Data}} & \multicolumn{1}{c}{\textbf{Noise}} \\
        \hline
        \colorbox{Canary Yellow}{Neutral} & \textbf{{[}B{]}} {\bng Aenk idn \colorbox{pink}{{O}eyT} \colorbox{Light Orange}{kra{I}echn} saeym bha{I}}\newline\textbf{{[}E{]}} \textit{You kept me waiting for several days brother Sayem.} & \colorbox{pink}{Mixed}\newline \colorbox{pink}{Language}\newline\colorbox{Light Orange}{Local Word} \\
        \hline
        \colorbox{Emerald Green}{Positive} & \textbf{{[}B{]}} {\bng Aaim maejh maejh Ja{I}\colorbox{Light Blue}{\textcolor{Light Blue}{-}} khub mangt khabar}\newline\textbf{{[}E{]}} \textit{I occasionally visit, and the food is of high quality.} & \colorbox{Light Blue}{Punctuation}\newline\colorbox{Light Blue}{Error} \\
        \hline
        \colorbox{Salmon}{Negative} & \textbf{{[}B{]}} {\bng bha{I} \colorbox{Buff}{dyaker} khabar nSh/T \colorbox{lightgray}{krebn/na}}\newline\textbf{{[}E{]}} \textit{Please don't waste food brother.} & \colorbox{Buff}{Spacing}\newline\colorbox{Buff}{Error}\newline \colorbox{lightgray}{Spelling}\newline\colorbox{lightgray}{Error} \\
        \hline
    \end{tabular}
    \end{adjustbox}
    \caption{Few examples from our \textbf{NC-SentNoB} dataset with sentiment on the leftmost column and noise types on the rightmost column. \textbf{B} represents the original text in Bangla and \textbf{E} represents the corresponding English translation.}
    \label{tab:samples}
\end{table}
With the recent emergence of pre-trained language models (PLMs) \citep{devlin2018bert, liu2019roberta, he2020deberta, raffel2020exploring, xue2020mt5}, there has been a notable enhancement in the sentiment analysis task. However, when confronted with increased textual noise, the performance of PLMs drops drastically (around 50\%), primarily due to the inability of the tokenizer to handle misspelled words \citep{srivastava2020noisy}. This issue is less pronounced in English, where most typing tools and applications offer robust auto-correction systems. However, Bangla, despite being the seventh most spoken language with a minimum of 272.7 million speakers \citep{wiki:benlan}, faces significant challenges due to the absence of an effective auto-correction system in digital devices and software. As a result, a considerable amount of text shared on social media platforms often exhibits diverse forms of noise, including informal language, regional words, spelling errors, typographic errors, punctuation errors, coined words, embedded metadata, a mixture of two or more languages (code-mixed text), grammatical mistakes and so forth \citep{srivastava2020noisy}. For example, the sentence "{\bng na , mu{I} {O}ega ikchu k{I} na{I} , du{h}khu pa{I}eb}" (\textit{English: No, I did not tell them anything, they will get sad}) incorporates regional words like "{\bng mu{I}}" ("{\bng Aaim}", \textit{I}), "{\bng {O}ega}" ("{\bng {O}edr}", \textit{them}), "{\bng k{I}}" ("{\bng bil}", \textit{tell}), "{\bng pa{I}eb}" ("{\bng paeb}", \textit{get}), alongside a spelling error "{\bng du{h}khu}" ("{\bng du{h}kh}", \textit{sad}).\par

Recent investigations into Bangla sentiment analysis have primarily focused on Bangla texts, Romanized Bangla texts \citep{hassan2016sentiment}, and social media comments \citep{chakraborty2022sentiment}. However, there is a notable scarcity of research specifically addressing noisy Bangla texts, and the available datasets for such studies are limited. To address this gap, the \textbf{SentNoB} dataset \citep{islam2021sentnob} has been recently introduced, aiming to tackle challenges associated with sentiment analysis in noisy Bangla texts. Nevertheless, it is worth noting that this dataset lacks annotations for noise types present in the noisy texts and does not incorporate any noise reduction methods. The presence of noise significantly impacts the performance of models compared to their performance on noiseless text, which indicates a potential area for further research. To address these issues, we have made the following contributions:
\begin{itemize}
    \item We present a dataset named \textbf{NC-SentNoB} (\textbf{N}oise \textbf{C}lassification on \textbf{SentNoB} dataset), designed for the identification of ten distinct types of noise found in approximately 15K noisy Bangla texts. Few sample instances are provided in Table \ref{tab:samples}.
    \item We employ machine learning, deep learning and fine-tune pre-trained transformer models to identify noise types in noisy Bangla texts (a multi-label classification task) and to perform sentiment analysis on both noisy and noise-reduced texts (a multi-class classification task).
    \item We conduct experiments with various techniques to reduce noise from Bangla texts including spell correction, back translation, paraphrasing and masking. To assess their effectiveness, we compare the performance of these methods against a set of 1000 random, noisy texts that have been manually corrected by annotators.
    \item We have made our dataset and codes openly accessible for further research in this field.
\end{itemize}

\section{Related Works}
\citet{haque2023multi} integrated 42,036 samples from two publicly available Bangla datasets, achieving the highest accuracy (85.8\%) in multi-class sentiment analysis with their proposed C-LSTM. \citet{islam2020sentiment} introduced two manually tagged Bangla datasets, achieving 71\% accuracy for binary classification and 60\% for multi-class classification using BERT with GRU. \citet{bhowmick2021sentiment} outperformed the baseline model proposed by \citet{islam2020sentiment}, attaining a 95\% accuracy on binary classification by fine-tuning \verb|m-BERT| and \verb|XLM-RoBERTa|. \citet{samia2022aspect} utilized \verb|BERT|, \verb|BiLSTM|, and \verb|LSTM| for aspect-based sentiment analysis, where \verb|BERT| performed best by achieving 95\% in aspect detection and 77\% sentiment classification. \citet{hasan2023natural} fine-tuned transformer models where \verb|BanglaBERT| surpassed other models with 86\% accuracy and a macro F1-score of 0.82 in multi-class setting.

Bangla sentiment analysis has also been extended to address the challenges of noisy social media texts. One of the notable contributions is SentNoB, a dataset of over 15,000 social media comments developed by \citet{islam2021sentnob}. It was benchmarked by SVM with lexical features, neural networks, and pre-trained language models. The best micro-averaged F1-Score (0.646) was achieved by \verb|SVM| with word and character n-grams. \citet{hoq2021sentiment} added Twitter data to SentNoB and got 87.31\% accuracy with multi-layer perceptrons. \citet{islam2023sentigold} developed SentiGOLD, which is a balanced Bangla sentiment dataset consisting of 70,000 entries with five classes which utilized SentNoB for cross-dataset evaluation. It was benchmarked by \verb|BiLSTM|, \verb|HAN|, \verb|BiLSTM|, \verb|CNN with attention| and \verb|BanglaBERT|. The best macro F1-Score (0.62) was achieved by fine-tuning \verb|BanglaBERT|, which also got an F1-Score (0.61) on SentNoB during cross-dataset testing.

As for the correction of noisy texts, \citet{koyama2021comparison} performed a comparative analysis of grammatical error correction using back-translation models. It was observed that the transformer-based model achieved the highest score on the CoNLL-2014 dataset \citep{ng-etal-2014-conll}. \citet{sun2019contextual} employed a \verb|BERT|-based masked language modeling for contextual noise reduction. This method involves sequentially masking and correcting each word in a sentence, starting from the left. They found that this noise reduction method significantly enhances performance in applications such as neural machine translation, natural language interfaces, and paraphrase detection in noisy texts.

\section{Noise Identification}
In this section, we first manually annotate all the instances from \textbf{SentNoB} dataset, categorizing them into \textbf{ten} separate noise categories. A single instance may fall into multiple noise categories. Then, we outline the process of noise identification, where the objective is to determine the type of noise present in a given noisy Bangla text. This task is framed as a multi-label classification task.

\subsection{Existing Dataset}
The \textbf{SentNoB} dataset \citep{islam2021sentnob} has a total of 15,728 noisy Bangla texts. While the dataset offers a collection of noisy Bangla texts, it lacks information regarding the specific types of noise present in these texts. The dataset is partitioned into three subsets: train (80\%), test (10\%), and validation (10\%). Each text is categorized into one of three labels: \textit{positive}, \textit{neutral}, and \textit{negative}. These labels represent the sentiment or tone expressed in each text.

\subsection{Dataset Development}
To the best of our knowledge, there is currently no dataset specifically designed for the purpose of identifying noise in Bangla texts. To address this gap, we expanded the SentNoB dataset to create a noise identification dataset named \textbf{NC-SentNoB} (\textbf{N}oise \textbf{C}lassification on \textbf{SentNoB} dataset), encompassing a total of 15,176 noisy texts. In the process, we eliminated 552 duplicate values present in the original dataset to enhance data integrity. We maintained the train-validation-test splitting ratio of the original dataset and the distribution of data in each partition is detailed in Table \ref{tab:SentNoB_Distribution}.

\begin{table}[h]
\centering
\resizebox{0.8\columnwidth}{!}{
\begin{tabular}{lrrr}
\hline
           & \textbf{Neutral} & \textbf{Positive} & \textbf{Negative} \\ \hline
\textbf{Train}      & 2,767     & 4,948     & 4,318    \\
\textbf{Test}       & 361      & 650      & 570     \\ 
\textbf{Validation}  & 354      & 621      & 587     \\ \hline
\textbf{Total}      & 3,482      & 6,219       & 5,475 \\ \hline
\end{tabular}
}
\caption{Data distribution in each partition.}\label{tab:SentNoB_Distribution}
\end{table}

\subsection{Annotation}
The primary idea behind developing the NC-SentNoB dataset was to categorize the noises available in the dataset. To do this, the authors thoroughly investigated the SentNoB dataset, determined ten categories, and defined rules for each noise type as the annotation guidelines. The details of each noise category are presented in Appendix \ref{ap1}. We first invited seven native Bangla speakers to assist us with the annotating process. Next, we asked each participant to label 50 samples, from which we determined their trustworthiness score \citep{price2020six}. We used 10 samples out of the 50 as control samples and discovered that only four participants achieved the 90\% trustworthiness score threshold. The degree of agreement across annotators is calculated using Fleiss' kappa score \citep{fleiss1971measuring} to maintain the quality of the annotation. After computing the scores for four independent annotators, we found a reliable score of 0.69, indicating a substantial degree of agreement. 

\begin{figure}[h]
    \centering
    \resizebox{\columnwidth}{!}{
    \includegraphics{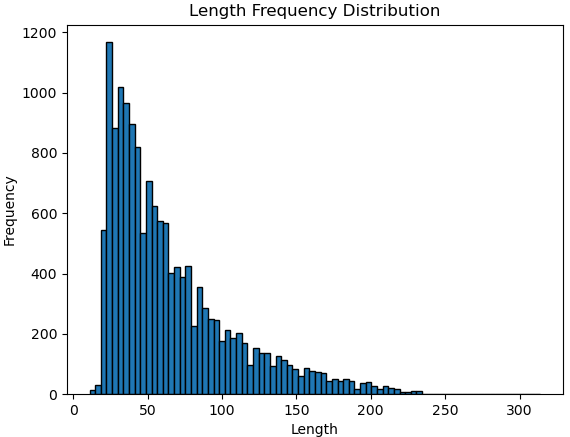}
    }
    \caption{Length-Frequency distribution of Texts.}
    \label{fig:LenFreq}
\end{figure}

\subsection{Dataset Statistics}
\begin{figure}[h]
    \centering
    \resizebox{\columnwidth}{!}{
    \includegraphics{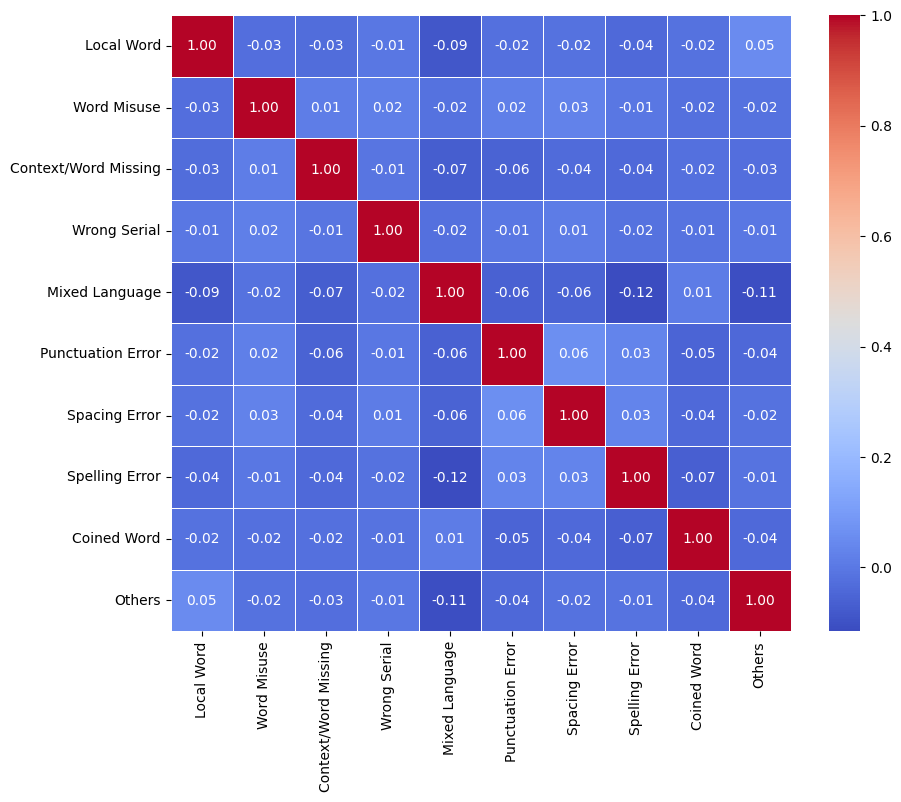}
    }
    \caption{Heatmap of correlation coefficients among different noise types in \textbf{NC-SentNoB}.}
    \label{fig:corr_heatmap}
\end{figure}

It is evident from Table \ref{tab:SentNoB_Distribution} that the dataset is imbalanced, with the number of texts in the \textit{neutral} category significantly lower than those in both the \textit{positive} and \textit{negative} categories.
\begin{table}[h]
\begin{adjustbox}{width=\columnwidth, center}
\begin{tabular}{lrc}
\hline
\multicolumn{1}{c}{\textbf{Class}} & \multicolumn{1}{c}{\textbf{Instances}} & \textbf{\#Word/Instance} \\ \hline

\textbf{Local Word}                     & 2,084 (0.136\%)                        & 16.05                    \\
\textbf{Word Misuse}               & 661 (0.043\%)                          & 18.55                    \\
\textbf{Context/Word Missing}      & 550 (0.036\%)                          & 13.19                    \\
\textbf{Wrong Serial}              & 69 (0.005\%)                           & 15.30                    \\
\textbf{Mixed Language}            & 6,267 (0.410\%)                        & 17.91                    \\
\textbf{Punctuation Error}               & 5,988 (0.391\%)                        & 17.25                    \\
\textbf{Spacing Error}                   & 2,456 (0.161\%)                        & 18.78                    \\
\textbf{Spelling Error}                  & 5,817 (0.380\%)                        & 17.30                    \\
\textbf{Coined Word}          & 549 (0.036\%)                          & 15.45                    \\ 
\textbf{Others}                     & 1,263 (0.083\%)                        & 16.52                    \\ \hline
\end{tabular}
\end{adjustbox}
\caption{Statistics of \textbf{NC-SentNoB} per noise class.}\label{tab:noise_stat}
\end{table}
In addition to the class imbalance, the dataset also exhibits a wide variation in the length of the texts. On an average, the texts have a length of 66 characters. The longest text is 314 characters, while the shortest text is only 11 characters long. Figure \ref{fig:LenFreq} shows the length frequency distribution of the texts over the whole dataset. Table \ref{tab:noise_stat} shows the statistics of different types of noise we found. This provides an insight into the most common noise of Bangla texts found on the dataset. The table shows that \emph{Mixed Language} is the most common noise type, \emph{Spelling Error} is the second most common, and \emph{Wrong Serial} is the least common. Figure \ref{fig:corr_heatmap} indicates low correlation coefficients, suggesting a minimal linear association between noise categories. Notably, \emph{Mixed Language} and \emph{Spelling Error} have the least correlation at -0.12, implying a slight inverse relationship between these two types. This indicates if a sentence in the dataset contains an error of \emph{Mixed Language}, it has a higher possibility of not having any \emph{Spelling Error} and vice versa.

\subsection{Baselines}
For noise identification, we implemented Support Vector Machine (\verb|SVM|) \citep{cortes1995support} (utilizing both character and word n-gram features), Bidirectional Long Short Term Memory (\verb|BiLSTM|) \citep{hochreiter1997long} network, and fine-tuned the pre-trained \verb|Bangla-BERT-Base| \citep{Sagor_2020} model. The descriptions of the models can be found in Appendix \ref{apA}. The rationale behind the classification is to develop an automatic text pre-processing step that identifies different types of noise present in Bangla texts. We firmly believe that this pre-processing step will play a vital role in addressing challenges associated with noisy Bangla texts by aiding in the development of noise specific reduction methods.

\subsection{Experimental Setup}
\verb|SVM| was implemented with a regularization parameter of 1. As for \verb|BiLSTM| and \verb|Bangla-BERT-Base|, Binary Cross-Entropy Loss was used. Both models were trained using the AdamW optimizer, with a learning rate of $1e-6$ for \verb|BiLSTM| and $1e-5$ for \verb|Bangla-BERT-Base|. The batch sizes were set at 256 for \verb|BiLSTM| and 128 for \verb|Bangla-BERT-Base|.

\subsection{Results \& Analysis}
Table \ref{scon} presents the performance comparison of the implemented models on noise identification. \verb|Bangla-BERT-Base| achieves the highest micro F1-score at 0.62, while \verb|SVM| with character-level features secures the second-best score of 0.57. However, \verb|BiLSTM| has the lowest micro F1-score of 0.24.
\begin{table}[h]
\centering
\resizebox{0.9\columnwidth}{!}{
\begin{tabular}{ l c c c }
\hline
    
\multicolumn{1}{c}{\textbf{Model}} & \multicolumn{1}{c}{\textbf{Precision}} & \multicolumn{1}{c}{\textbf{Recall}} & \textbf{F1-Score} \\ \hline
\textbf{SVM (C)}    & 0.76               & 0.45            & 0.57              \\ 
\textbf{SVM (W)} & 0.64   &   0.38   &   0.48\\
\textbf{SVM (C + W)} & 0.75   &   0.45  &    0.56\\
\textbf{Bi-LSTM}    & 0.36               & 0.18            & 0.24              \\ 
\textbf{Bangla-BERT-Base} & 0.73               & 0.54            & \textbf{0.62}              \\ \hline
\end{tabular}
}
\caption{Performance comparison of different models on noise identification. \textbf{C} represents character level n-gram and \textbf{W} represents word level n-gram.}\label{scon}
\end{table}
The comparison between \verb|SVM| with character-level features and \verb|SVM| with word-level features shows that the former attains a higher score. This suggests that character-level information is more crucial for noise identification. Implementing a similar character-level approach in neural network models and fine-tuning other pre-trained language models may improve the noise identification performance which we leave open for future work. Table \ref{tab:cw-perf-bert} illustrates the performance of \verb|Bangla-BERT-Base| on each type of noise. It can be seen that the model fails to classify instances of the \textit{Wrong Serial} type. This is primarily due to the low amount of data available for this specific class in the dataset.

\begin{table}[h!]
    \centering
    \resizebox{\columnwidth}{!}{
    \begin{tabular}{lccc}
    \hline
    \multicolumn{1}{c}{\textbf{Class}} & \multicolumn{1}{c}{\textbf{Precision}} & \multicolumn{1}{c}{\textbf{Recall}} & \textbf{F1-Score} \\ \hline
\textbf{Local Word}                & 0.46 & 0.49 & 0.47 \\
\textbf{Word Misuse}          & 0.65 & 0.16 & 0.25 \\
\textbf{Context/Word Missing} & 0.33 & 0.06 & 0.10 \\
\textbf{Wrong Serial}         & 0.00 & 0.00 & 0.00 \\
\textbf{Mixed Language}       & 0.75 & 0.85 & 0.80 \\
\textbf{Punctuation Error}          & 0.83 & 0.54 & 0.65 \\
\textbf{Spacing Error}              & 0.86 & 0.21 & 0.33 \\
\textbf{Spelling Error}             & 0.64 & 0.55 & 0.59 \\
\textbf{Coined Word}         & 0.82 & 0.89 & 0.86 \\ 
\textbf{Others}               & 0.76 & 0.76 & 0.76 \\ \hline    
\end{tabular}
}
    \caption{Class-wise performance of \textbf{Bangla-BERT-Base} on noise identification task.}
    \label{tab:cw-perf-bert}
\end{table}

\section{Sentiment Analysis}
In this section, we outline the methodology employed for conducting sentiment analysis on the NC-SentNoB dataset. We employ a cost-sensitive learning objective to fine-tune seven pre-trained transformer models for the sentiment analysis task. We conduct two distinct experiments: the first involves fine-tuning transformers on the noisy texts, while the second entails fine-tuning transformers after reducing noise from the original noisy texts.
\subsection{Baselines}
We utilized seven publicly available pre-trained transformer models: \verb|Bangla-Bert-Base| \citep{Sagor_2020}, \verb|BanglaBERT| \citep{bhattacharjee-etal-2022-banglabert}, \verb|BanglaBERT Large| \citep{bhattacharjee-etal-2022-banglabert}, \verb|SahajBERT|\footnote{\url{https://huggingface.co/neuropark/sahajBERT}}, \verb|Bangla-Electra|\footnote{\url{https://huggingface.co/monsoon-nlp/bangla-electra}}, \verb|MuRIL| \citep{khanuja2021muril}. The descriptions of the models can be found in Appendix \ref{apA}.

\subsection{Cost-sensitive Learning}
Cost-sensitive learning \citep{elkan2001foundations} is a process of training where we can make the model prioritize samples from the minority class above those from the majority class by suggesting a manually established weight for every class label in the cost function that is being minimized. We adopted this method in the sentiment analysis task. In order to provide a more equitable and balanced model performance, we tried imposing larger costs to the classes that are in the minority in numbers due to the imbalance scenario in the NC-SentNoB dataset, as seen in Table \ref{tab:SentNoB_Distribution}. This was accomplished by providing class weights to the Cross-Entropy loss function used to train the models.

\subsection{Experimental Setup}
Cost-sensitive learning was incorporated by using class weights as a cost matrix into the Cross-Entropy loss function. The class weights were set at 1.4496 for \textit{neutral}, 0.8106 for \textit{positive}, and 0.9289 for \textit{negative} classes. For fine-tuning, the AdamW optimizer was used with a learning rate of $1e-5$, betas set at (0.9, 0.9999), an epsilon value of $1e-9$, and a weight decay of 0.08. Due to resource constraints, batch size was set to 48 for \verb|sahajBERT|, 32 for \verb|BanglaBERT Large|, and 128 for the rest of the models.

\begin{table}[h]
    \centering
    \resizebox{\columnwidth}{!}{
    \begin{tabular}{lccc}
        \hline
        \multicolumn{1}{c}{\textbf{Model}} & \multicolumn{1}{c}{\textbf{Precision}} & \multicolumn{1}{c}{\textbf{Recall}} & \textbf{F1-Score} \\
        \hline
        \textbf{Bangla-BERT-Base} & 0.72 & 0.72 & 0.72 \\
        \textbf{BanglaBERT} & 0.75 & 0.75 & \textbf{0.75} \\
        \textbf{BanglaBERT Large} & 0.74 & 0.74 & 0.74 \\
        \textbf{BanglaBERT Generator} & 0.72 & 0.72 & 0.72 \\
        \textbf{sahajBERT} & 0.72 & 0.72 & 0.72 \\
        \textbf{Bangla-Electra} & 0.68 & 0.68 & 0.68 \\
        \textbf{MuRIL} & 0.73 & 0.73 & 0.73 \\
        \hline
    \end{tabular}
    }
    \caption{Performance of sentiment analysis models fine-tuned on noisy texts.}
    \label{tab:perf_noisy}
\end{table}

\subsection{Experiment with noise}\label{Sec:Noise}
Table \ref{tab:perf_noisy} illustrates the performance comparison of the seven fine-tuned models. \verb|BanglaBERT| yields the highest scores across all evaluation metrics with a micro F1-score of $0.75$. This result outperforms the highest micro F1-Score of 0.6461 with \verb|SVM| previously reported by \citet{islam2021sentnob}. It is also noteworthy that all other models except \verb|Bangla-Electra| have demonstrated results that are somewhat comparable with ranges between 0.72 and 0.75 in terms of micro F1-score. 

\subsection{Experiment by reducing noise} \label{Sec:wNoise}
In this experiment, we first outline the noise reduction strategies utilized prior to sentiment analysis. We then randomly select 1000 noisy texts and manually correct them. We use these 1000 manually corrected texts as ground truth for measuring the performance of the noise reduction methods in terms of semantic similarity. To assess performance, we employ various established evaluation metrics.
\begin{table}[h]
\centering
\resizebox{0.7\columnwidth}{!}{
\begin{tabular}{lr}
\hline
\multicolumn{1}{c}{\textbf{Class}} & \multicolumn{1}{c}{\textbf{Instances}} \\ \hline
\textbf{Local Word}                & 132 (13.2\%)                           \\
\textbf{Word Misuse}               & 32 (03.2\%)                            \\
\textbf{Context/Word Missing}      & 39 (03.9\%)                            \\
\textbf{Wrong Serial}              & 4 (00.4\%)                             \\
\textbf{Mixed Language}            & 416 (41.6\%)                           \\
\textbf{Punctuation Error}         & 323 (32.3\%)                           \\
\textbf{Spacing Error}             & 133 (13.3\%)                           \\
\textbf{Spelling Error}            & 376 (37.6\%)                           \\
\textbf{Coined Word}              & 33 (03.3\%)                            \\
\textbf{Others}                    & 92 (09.2\%)                            \\ \hline
\end{tabular}
}
\caption{Statistics of noise types on manually corrected 1000 data.}
\label{tab:stats_1000}
\end{table}
\begin{table*}[t]
\centering
\resizebox{\textwidth}{!}{
\begin{tabular}{l c c c c c c ccc c }
\hline
\multicolumn{1}{c}{\multirow{2}{*}{}}                               & \multirow{2}{*}{\textbf{BLEU}} & \multirow{2}{*}{\textbf{ROUGE-L}} & \multirow{2}{*}{\textbf{\begin{tabular}[c]{@{}c@{}}BERT\\ Score\end{tabular}}} & \multirow{2}{*}{\textbf{\begin{tabular}[c]{@{}c@{}}SBERT\\ Score\end{tabular}}} & \multirow{2}{*}{\textbf{BSTS}} & \multirow{2}{*}{\textbf{\begin{tabular}[c]{@{}c@{}}BERT\\ -iBLEU\end{tabular}}} & \multicolumn{3}{c}{\textbf{Word Coverage}}                                                                                                      & \multirow{2}{*}{\textbf{\begin{tabular}[c]{@{}c@{}}Human\\ Evaluation\\(\%)\end{tabular}}} \\ \cline{8-10}
\multicolumn{1}{c}{}                                                &                                &                                   &                                &                                                                                         &                                &                                                                                 & \multicolumn{1}{c}{\textbf{Word2Vec}} & \multicolumn{1}{c}{\textbf{FastText}} & \textbf{\begin{tabular}[c]{@{}c@{}}Bangla\\ BERT\end{tabular}} &                                                                                      \\ \hline
\textbf{Noisy Text}                                                   & \textbf{65.77}                & \textbf{79.71}                    & \textbf{93.21}                 & \textbf{88.32}                                                                          & \textbf{93.67}                          & 51.65                                                                           & \multicolumn{1}{c}{75.54}             & \multicolumn{1}{c}{82.92}             & 71.26                                                          & X                                                                                       \\ \hline
\textbf{\begin{tabular}[c]{@{}l@{}}Google\\ Translate\end{tabular}}   & 21.55                         & 39.46                             & 84.72                          & 81.04                                                                                   & 84.28                          & \textbf{80.93}                                                                  &  \multicolumn{1}{c}{87.52}             & \multicolumn{1}{c}{\textbf{89.01}}    & 84.86                                                          & \textbf{37.90}                                                                                      \\ \hline
\textbf{\begin{tabular}[c]{@{}l@{}}BanglaT5\\ Translate\end{tabular}} & 16.57                         & 32.09                             & 81.30                          & 75.27                                                                                   & 82.15                          & 80.12                                                                           & \multicolumn{1}{c}{\textbf{89.01}}    & \multicolumn{1}{c}{87.52}             & 85.66                                                          & 21.10                                                                                     \\ \hline
\textbf{\begin{tabular}[c]{@{}l@{}}Spell \\ Correction (SC)\end{tabular}}  & 61.17                         & 77.35                             & 92.29                          & 87.86                                                                                   & 92.94                 & 56.50                                                                           & \multicolumn{1}{c}{82.72}             & \multicolumn{1}{c}{88.51}             & 80.76                                                          & 35.80                                                                                     \\ \hline
\textbf{\begin{tabular}[c]{@{}l@{}}SC +\\ Paraphrase\end{tabular}}    & 20.35                         & 36.44                             & 83.32                          & 74.15                                                                                   & 85.60                          & 80.63                                                                           & \multicolumn{1}{c}{86.79}             & \multicolumn{1}{c}{86.79}             & 83.89                                                          & 20.80                                                                                      \\ \hline
\textbf{\begin{tabular}[c]{@{}l@{}}MLM \\(OOV)\end{tabular}}                                                     & 60.99                         & 76.44                             & 90.72                          & 86.90                                                                                    & 91.82                          & 56.60                                                                           & \multicolumn{1}{c}{88.51}             & \multicolumn{1}{c}{82.27}             & 87.18                                                 & 26.80                                                                                      \\ \hline
\textbf{\begin{tabular}[c]{@{}l@{}}MLM \\(Random)\end{tabular}}                                                     & 44.17                         & 70.00                             & 90.76                          & 85.26                                                                                    & 93.45                          & 68.93                                                                           & \multicolumn{1}{c}{86.41}             & \multicolumn{1}{c}{88.35}             & \textbf{93.20}                                                 & 10.40                                                                                      \\ \hline
\end{tabular}
}
\caption{Performance comparison of different noise reduction methods}\label{den}
\end{table*}
\subsubsection{Process of Noise Reduction}
Complete elimination of noise from the noisy texts is impossible. However, our aim is to minimize noise to the greatest extent possible. This section details four distinct methods for reducing noise in noisy texts: back-translation, spelling correction, paraphrasing and replacing out of vocabulary (OOV) words with predictions generated by a masked language model (MLM). Additional details about the employed methods can be found in Appendix \ref{apA}. \\
\noindent\textbf{(a) Back-translation.} 
Back-translation serves as a method to correct various errors within a sentence. As pre-trained models have been trained on extensive corpora of noiseless sentences, they can generate a noiseless translated sentence when presented with a noisy sentence as input. Also, translating that sentence back into the original language may result in a corrected version. For this study, all input texts were initially translated into English and then into Bangla using back-translation. Two models were chosen for this purpose: Google Translate, a web service employing an RNN-based model and \verb|BanglaT5| models pre-trained on the BanglaNMT English-Bangla and BanglaNMT Bangla-English dataset \citep{bhattacharjee2022banglanlg}.\\
\textbf{(b) Spelling Correction.} For the noisy texts we are working with, correcting spelling errors can be a beneficial process as spelling errors can affect the tokenization process. To address this issue, we implemented a spell correction algorithm based on \verb|Soundex| and \verb|Levenshtein| distance. This algorithm replaces misspelled words with the closest matching words found in the Bangla dictionary\footnote{\url{https://github.com/MinhasKamal/BanglaDictionary}}. However, as it is not context-based, there are instances where it fails to correct all spellings and may even introduce out-of-context words in the sentence.\\
\textbf{(c) Paraphrasing.} Paraphrasing involves changing the words of a sentence without altering its meaning. Similar to translation models, paraphrasing models have the potential to provide a noiseless paraphrased output when given a noisy input. For this study, we used the \verb|BanglaT5| model pre-trained on the Bangla Paraphrase dataset \citep{akil2022banglaparaphrase}. We observed the performance of the \verb|BanglaT5| paraphrase model on some randomly selected noisy texts from our dataset and found that the model performs poorly when the input data contains misspelled words. To address this issue, we used the spelling corrector algorithm prior providing input to the model.\\
\textbf{(d) Mask Prediction.} To improve the quality of noisy texts and address out-of-vocabulary words, we replaced OOV words with <MASK> and used the predictions generated by a Masked Language Model (MLM). We also implemented random masking for replacement with each word having a 20\% possibility of getting replaced by the MLM model. For both cases, we used \verb|BanglaBERT Generator| \citep{kowsher2022bangla} model.

\subsubsection{Evaluation of Noise Reduction}
We first use several well-known metrics to quantify the performance of the noise reduction techniques. The evaluation is performed based on 1000 manually corrected texts. The first four authors individually corrected 250 texts each, while the last two authors verified corrections for 500 texts each. We then compare and analyze the performance of the noise reduction methods.\\
\textbf{Evaluation Metrics.} To evaluate the noise reduction methods, we employed a range of metrics including BLEU, ROUGE-L, BERTScore, SBERT Score, BSTS, BERT-iBLEU, and Word Coverage (utilizing Word2Vec, FastText, and \verb|Bangla-BERT-Base|). Additionally, we conducted human evaluations of the noise reduced sentences by native Bangla speakers. The detailed descriptions of the evaluation metrics along with the human evaluation procedure are presented in Appendix \ref{App:denoise}.\\
\textbf{Noise Reduction Performance.} From the data presented in Table \ref{den}, it can be seen that the original noisy texts scored highest on BLEU and ROUGE-L, which is unsurprising since the ground truth sentences contain nearly identical words. This observation is further supported by the spell-corrected sentences, which also achieve a similar score due to having nearly identical words.
\begin{table}[h!]
    \centering
    \resizebox{0.9\columnwidth}{!}{
    \begin{tabular}{m{0.05\columnwidth}|m{0.85\columnwidth}}
    \hline
    \rotatebox{90}{\parbox{3cm}{\textbf{Before reduction}}} & 
    \textbf{[N]} {\bng Aapin eta Hat \textcolor{red}{dhya} bhuel egeln bha{I} } \newline
    \textbf{[C]} {\bng Aapin eta Hat edhaya \textcolor{ForestGreen}{bhuel} egeln bha{I} } \newline
    \textbf{[E]} Brother you forgot to wash your hands.
    \\ \hline
    \rotatebox{90}{\parbox{3cm}{\textbf{After reduction}}} & 
    \textbf{[S]} {\bng Aapin eta Hat \textcolor{red}{dya} bhuel egeln bha{I} } \newline
    \textbf{[SP]} {\bng tuim etamar \textcolor{red}{Hat}-payer \textcolor{red}{dya} bhuel egch, bha{I}. } \newline 
    \textbf{[TG]} {\bng bha{I} Aapin Hat \textcolor{ForestGreen}{dhuet} bhuel egechn } \newline
    \textbf{[TM]} {\bng tuim Hat \textcolor{red}{dhret} bhuel egecha } \newline
    \textbf{[MO]} {\bng Aapin eta Hat \textcolor{red}{Haraet} bhuel egeln bha{I} } \newline
    \textbf{[MR]} {\bng Aapin eta Hat \textcolor{red}{dhret} bhuel egeln bha{I} }
    \\ \hline
    \end{tabular}
    }
    \caption{Input and output of a single noisy text by the noise reduction methods. \textbf{N} denotes the original noisy text, \textbf{C} indicates the corrected text, and \textbf{E} represents English translation of the corrected text. \textbf{S}, \textbf{SP}, \textbf{TG}, \textbf{TM}, \textbf{MO}, and \textbf{MR} represent outputs of spelling correction, paraphrasing with spelling correction, back-translation using Google Translate, back-translation with T5 models, masked language modeling for out-of-vocabulary words, and random masked language modeling respectively. For each sentence, noisy words are marked with \textcolor{red}{Red} color, and noise reduced words are marked with \textcolor{ForestGreen}{Green} color.}
    \label{tab:sample_denoise}
\end{table}
Similarly, for BERTScore, SBERT Score, and BSTS, the scores are higher for noisy texts. This is primarily because of the nature of textual embeddings and the tokenization method used. As mentioned earlier, \verb|BERT| uses WordPiece tokenization, which can result in identical words having the same token. Therefore, when comparing noisy texts with their corresponding ground truth sentences, many tokens are likely to match perfectly, leading to higher cosine similarity scores. However, although not having the highest score, back-translation, paraphrasing, and mask prediction methods score above 80\% in both BERTScore and BSTS, implying that they are semantically similar and the meaning of the sentences have not changed drastically. BERT-iBLEU score accounts for the presence of textually similar words by applying penalization while emphasizing semantic meaning, leading to Google Translate achieving the highest score in this metric. Moreover, the word coverage results show different methods scoring the highest instead of noisy texts. This is due to the generated words or sentences from these models having a higher possibility of being noiseless words from their respective vocabularies. 
All of the scores are based on the textual similarity of the ground truths and noise reduced sentences. Thus, we relied on human evaluation to select the best noise reduction method where 4 native Bangla speakers evaluated the sentences and discovered that the back-translation method utilizing Google Translate API was the most reliable in terms of maintaining contextual meaning. The input and output of each noise reduction method for a single noisy text are shown in table \ref{tab:sample_denoise}. Except for back-translation using Google Translate, all methods fail to rectify the spelling problem in the input. Most approaches change the meaning of the sentence by changing the noisy word.

\subsubsection{Results \& Analysis}
We prioritized the human evaluation score based on the results of Table \ref{den} and used back-translated data obtained from Google Translate to execute the sentiment analysis task by fine-tuning seven pre-trained transformer models. We applied the same noise reduction method on both the test and validation sets. We compared the sentiment analysis performance of the models fine-tuned on noisy and noiseless data presented in Tables \ref{tab:perf_noisy} and \ref{tab:perf_denoisy}. 
\begin{table}[h]
    \centering
    \resizebox{\columnwidth}{!}{
    \begin{tabular}{lccc}
        \hline
         \multicolumn{1}{c}{\textbf{Model}} & \multicolumn{1}{c}{\textbf{Precision}} & \multicolumn{1}{c}{\textbf{Recall}} & \textbf{F1-Score} \\
        \hline
        \textbf{Bangla-BERT-Base} & 0.69 & 0.69 & 0.69 \\
        \textbf{BanglaBERT} & 0.72 & 0.72 & 0.72 \\
        \textbf{BanglaBERT Large} & 0.73 & 0.73 & \textbf{0.73} \\
        \textbf{BanglaBERT Generator} & 0.70 & 0.70 & 0.70 \\
        \textbf{sahajBERT} & 0.70 & 0.70 & 0.70 \\
        \textbf{Bangla-Electra} & 0.66 & 0.66 & 0.66 \\
        \textbf{MuRIL} & 0.71 & 0.71 & 0.71 \\
        \hline
    \end{tabular}
    }
    \caption{Performance of sentiment analysis models fine-tuned on noise reduced texts (back-translation with google translate).}
    \label{tab:perf_denoisy}
\end{table}
From Table \ref{tab:perf_denoisy}, it can be seen that models fine-tuned on back-translated data only attain the highest F1-Score of $0.73$. This outcome remains consistent across all models evaluated during our experimentation. The model fine-tuned on noisy data outperformed the same model fine-tuned on back-translated data. The reason for this disparity of performance is that, while back-translation can mitigate some sources of noise, it can also introduce changes in the contextual meaning of the sentences (see Appendix \ref{ap2}). Because of this, it had a score of 37.90\% on human evaluation where our main priority of scoring was the contextual meaning of the sentence. We used the human evaluation score to achieve the best noise reduction strategy, although as shown in Table \ref{den}, other techniques scored well on several metrics as well. Nevertheless, it is worthwhile to explore alternative approaches beyond back-translation to determine whether a particular noise reduction method yields superior results in addressing specific types of noisy texts.
\begin{table}[h]
    \centering
    \resizebox{0.8\columnwidth}{!}{
    \begin{tabular}{lccc}
    \hline
         \multicolumn{1}{c}{\textbf{Class}} & \multicolumn{1}{c}{\textbf{Precision}} & \multicolumn{1}{c}{\textbf{Recall}} & \textbf{F1-Score} \\ \hline
         \textbf{Neutral} & 0.53 & 0.51 & 0.52 \\
         \textbf{Positive} & 0.77 & 0.77 & 0.77 \\
         \textbf{Negative} & 0.78 & 0.80 & 0.79 \\ \hline
         \textbf{Micro} & 0.73 & 0.73 & 0.73 \\
         \textbf{Macro} & 0.69 & 0.69 & 0.69 \\
         \textbf{Weighted} & 0.72 & 0.73 & 0.72 \\ \hline
    \end{tabular}
    }
    \caption{Class-wise performance of \textbf{BanglaBERT Large} on noise reduced texts (back-translation with google translate).}
    \label{tab:cw-perf-gt}
\end{table}
 Table \ref{tab:cw-perf-gt} illustrates the class-wise results of our best-performing model - \verb|BanglaBERT Large| on noise reduced data. It is clear from the table that the results are quite high for the positive and negative classes but the opposite for the neutral class. Few training data points might be the reason for this low performance in that particular class.

\section{Limitations and Future Works}
One obvious limitation is that none of the noise reduction methods we employed were able to correctly reduce noise from the noisy texts. As a result, fine-tuned models achieved a lower score in sentiment analysis than models fine-tuned on noisy texts. Another limitation is that we have not evaluated sentiment analysis by considering alternative noise reduction techniques other than back-translation by Google Translate. Although other noise reduction methods performed poorly in human evaluation, it would be interesting to study whether their performance in noise reduction correlates with the performance in sentiment analysis. Furthermore, the NC-SentNoB dataset contains only a very small number of \textit{Wrong Serial} data instances. Other categories such as \textit{Context/Word Missing}, \textit{Word Misuse}, and \textit{Coined Word} are also underrepresented. In future, we would like to increase the data in these categories to tackle data imbalance, which may potentially enhance the performance of the transformer models. In addition, to combat noise coming from spelling variation and dialectal differences, we plan to incorporate text normalization methods i.e. character-level spell correction models \citep{farra-etal-2014-generalized, Zaky2019AnLS} and character-level Neural Machine Translation (NMT) models \citep{lee-etal-2017-fully, edman2023character} for back-translation. We hypothesize that text normalization methods might be a viable solution due to their ability to comprehend context at character level. Finally, we will investigate noise-specific reduction techniques and report on the noise reduction approaches that demonstrate superior results in addressing particular types of noisy texts.

\section{Conclusion}
This study involves a comparison of various noise reduction techniques to assess their effectiveness in reducing noise within the NC-SentNoB dataset, which includes ten distinct types of noises. The results indicate that none of the noise reduction methods effectively reduce noise in the texts, leading to a lower F1-score compared to the sentiment analysis of noisy texts. This underscores the necessity for the development of noise-specific reduction techniques. We conducted a statistical analysis of our NC-SentNoB dataset and employed baseline models to identify the noises. However, the data imbalance adversely impacts the model performance suggesting potential enhancement upon addressing this imbalance.

\bibliography{anthology,custom}

\begin{thebibliography}{47}
\expandafter\ifx\csname natexlab\endcsname\relax\def\natexlab#1{#1}\fi

\bibitem[{Akil et~al.(2022)Akil, Sultana, Bhattacharjee, and Shahriyar}]{akil2022banglaparaphrase}
Ajwad Akil, Najrin Sultana, Abhik Bhattacharjee, and Rifat Shahriyar. 2022.
\newblock Banglaparaphrase: A high-quality bangla paraphrase dataset.
\newblock \emph{arXiv preprint arXiv:2210.05109}.

\bibitem[{Bhattacharjee et~al.(2022{\natexlab{a}})Bhattacharjee, Hasan, Ahmad, Mubasshir, Islam, Iqbal, Rahman, and Shahriyar}]{bhattacharjee-etal-2022-banglabert}
Abhik Bhattacharjee, Tahmid Hasan, Wasi Ahmad, Kazi~Samin Mubasshir, Md~Saiful Islam, Anindya Iqbal, M.~Sohel Rahman, and Rifat Shahriyar. 2022{\natexlab{a}}.
\newblock \href {https://doi.org/10.18653/v1/2022.findings-naacl.98} {{B}angla{BERT}: Language model pretraining and benchmarks for low-resource language understanding evaluation in {B}angla}.
\newblock In \emph{Findings of the Association for Computational Linguistics: NAACL 2022}, pages 1318--1327, Seattle, United States. Association for Computational Linguistics.

\bibitem[{Bhattacharjee et~al.(2022{\natexlab{b}})Bhattacharjee, Hasan, Ahmad, and Shahriyar}]{bhattacharjee2022banglanlg}
Abhik Bhattacharjee, Tahmid Hasan, Wasi~Uddin Ahmad, and Rifat Shahriyar. 2022{\natexlab{b}}.
\newblock \href {http://arxiv.org/abs/2205.11081} {Banglanlg: Benchmarks and resources for evaluating low-resource natural language generation in bangla}.
\newblock \emph{CoRR}, abs/2205.11081.

\bibitem[{Bhowmick and Jana(2021)}]{bhowmick2021sentiment}
Anirban Bhowmick and Abhik Jana. 2021.
\newblock Sentiment analysis for bengali using transformer based models.
\newblock In \emph{Proceedings of the 18th International Conference on Natural Language Processing (ICON)}, pages 481--486.

\bibitem[{Chakraborty et~al.(2022)Chakraborty, Nawar, and Chowdhury}]{chakraborty2022sentiment}
Partha Chakraborty, Farah Nawar, and Humayra~Afrin Chowdhury. 2022.
\newblock Sentiment analysis of bengali facebook data using classical and deep learning approaches.
\newblock In \emph{Innovation in Electrical Power Engineering, Communication, and Computing Technology: Proceedings of Second IEPCCT 2021}, pages 209--218. Springer.

\bibitem[{Clark et~al.(2020)Clark, Luong, Le, and Manning}]{clark2020electra}
Kevin Clark, Minh-Thang Luong, Quoc~V Le, and Christopher~D Manning. 2020.
\newblock Electra: Pre-training text encoders as discriminators rather than generators.
\newblock \emph{arXiv preprint arXiv:2003.10555}.

\bibitem[{Cortes and Vapnik(1995)}]{cortes1995support}
Corinna Cortes and Vladimir Vapnik. 1995.
\newblock Support-vector networks.
\newblock \emph{Machine learning}, 20:273--297.

\bibitem[{Devlin et~al.(2018{\natexlab{a}})Devlin, Chang, Lee, and Toutanova}]{devlin2018bert}
Jacob Devlin, Ming-Wei Chang, Kenton Lee, and Kristina Toutanova. 2018{\natexlab{a}}.
\newblock Bert: Pre-training of deep bidirectional transformers for language understanding.
\newblock \emph{arXiv preprint arXiv:1810.04805}.

\bibitem[{Devlin et~al.(2018{\natexlab{b}})Devlin, Chang, Lee, and Toutanova}]{DBLP:journals/corr/abs-1810-04805}
Jacob Devlin, Ming{-}Wei Chang, Kenton Lee, and Kristina Toutanova. 2018{\natexlab{b}}.
\newblock \href {http://arxiv.org/abs/1810.04805} {{BERT:} pre-training of deep bidirectional transformers for language understanding}.
\newblock \emph{CoRR}, abs/1810.04805.

\bibitem[{Edman et~al.(2023)Edman, Toral, and van Noord}]{edman2023character}
Lukas Edman, Antonio Toral, and Gertjan van Noord. 2023.
\newblock Are character-level translations worth the wait? an extensive comparison of character-and subword-level models for machine translation.
\newblock \emph{arXiv preprint arXiv:2302.14220}.

\bibitem[{Elkan(2001)}]{elkan2001foundations}
Charles Elkan. 2001.
\newblock The foundations of cost-sensitive learning.
\newblock In \emph{International joint conference on artificial intelligence}, volume~17, pages 973--978. Lawrence Erlbaum Associates Ltd.

\bibitem[{Farra et~al.(2014)Farra, Tomeh, Rozovskaya, and Habash}]{farra-etal-2014-generalized}
Noura Farra, Nadi Tomeh, Alla Rozovskaya, and Nizar Habash. 2014.
\newblock \href {https://doi.org/10.3115/v1/P14-2027} {Generalized character-level spelling error correction}.
\newblock In \emph{Proceedings of the 52nd Annual Meeting of the Association for Computational Linguistics (Volume 2: Short Papers)}, pages 161--167, Baltimore, Maryland. Association for Computational Linguistics.

\bibitem[{Fleiss(1971)}]{fleiss1971measuring}
Joseph~L Fleiss. 1971.
\newblock Measuring nominal scale agreement among many raters.
\newblock \emph{Psychological bulletin}, 76(5):378.

\bibitem[{Haque et~al.(2023)Haque, Islam, Tasneem, and Das}]{haque2023multi}
Rezaul Haque, Naimul Islam, Mayisha Tasneem, and Amit~Kumar Das. 2023.
\newblock Multi-class sentiment classification on bengali social media comments using machine learning.
\newblock \emph{International Journal of Cognitive Computing in Engineering}, 4:21--35.

\bibitem[{Hasan et~al.(2023)Hasan, Islam, Jahan, Meem, and Rahman}]{hasan2023natural}
Mahmud Hasan, Labiba Islam, Ismat Jahan, Sabrina~Mannan Meem, and Rashedur~M Rahman. 2023.
\newblock Natural language processing and sentiment analysis on bangla social media comments on russia--ukraine war using transformers.
\newblock \emph{Vietnam Journal of Computer Science}, pages 1--28.

\bibitem[{Hassan et~al.(2016)Hassan, Amin, Al~Azad, and Mohammed}]{hassan2016sentiment}
Asif Hassan, Mohammad~Rashedul Amin, Abul~Kalam Al~Azad, and Nabeel Mohammed. 2016.
\newblock Sentiment analysis on bangla and romanized bangla text using deep recurrent models.
\newblock In \emph{2016 International Workshop on Computational Intelligence (IWCI)}, pages 51--56. IEEE.

\bibitem[{He et~al.(2020)He, Liu, Gao, and Chen}]{he2020deberta}
Pengcheng He, Xiaodong Liu, Jianfeng Gao, and Weizhu Chen. 2020.
\newblock Deberta: Decoding-enhanced bert with disentangled attention.
\newblock \emph{arXiv preprint arXiv:2006.03654}.

\bibitem[{Hochreiter and Schmidhuber(1997)}]{hochreiter1997long}
Sepp Hochreiter and J{\"u}rgen Schmidhuber. 1997.
\newblock Long short-term memory.
\newblock \emph{Neural computation}, 9(8):1735--1780.

\bibitem[{Hoq et~al.(2021)Hoq, Haque, and Uddin}]{hoq2021sentiment}
Muntasir Hoq, Promila Haque, and Mohammed~Nazim Uddin. 2021.
\newblock Sentiment analysis of bangla language using deep learning approaches.
\newblock In \emph{International Conference on Computing Science, Communication and Security}, pages 140--151. Springer.

\bibitem[{Islam et~al.(2020)Islam, Islam, and Amin}]{islam2020sentiment}
Khondoker~Ittehadul Islam, Md~Saiful Islam, and Md~Ruhul Amin. 2020.
\newblock Sentiment analysis in bengali via transfer learning using multi-lingual bert.
\newblock In \emph{2020 23rd International Conference on Computer and Information Technology (ICCIT)}, pages 1--5. IEEE.

\bibitem[{Islam et~al.(2021)Islam, Kar, Islam, and Amin}]{islam2021sentnob}
Khondoker~Ittehadul Islam, Sudipta Kar, Md~Saiful Islam, and Mohammad~Ruhul Amin. 2021.
\newblock Sentnob: A dataset for analysing sentiment on noisy bangla texts.
\newblock In \emph{Findings of the Association for Computational Linguistics: EMNLP 2021}, pages 3265--3271.

\bibitem[{Islam et~al.(2023)Islam, Chowdhury, Khan, Hossain, Hossain, Rashid, Mohammed, and Amin}]{islam2023sentigold}
Md~Ekramul Islam, Labib Chowdhury, Faisal~Ahamed Khan, Shazzad Hossain, Sourave Hossain, Mohammad Mamun~Or Rashid, Nabeel Mohammed, and Mohammad~Ruhul Amin. 2023.
\newblock Sentigold: A large bangla gold standard multi-domain sentiment analysis dataset and its evaluation.
\newblock \emph{arXiv preprint arXiv:2306.06147}.

\bibitem[{Khanuja et~al.(2021)Khanuja, Bansal, Mehtani, Khosla, Dey, Gopalan, Margam, Aggarwal, Nagipogu, Dave, Gupta, Gali, Subramanian, and Talukdar}]{khanuja2021muril}
Simran Khanuja, Diksha Bansal, Sarvesh Mehtani, Savya Khosla, Atreyee Dey, Balaji Gopalan, Dilip~Kumar Margam, Pooja Aggarwal, Rajiv~Teja Nagipogu, Shachi Dave, Shruti Gupta, Subhash Chandra~Bose Gali, Vish Subramanian, and Partha Talukdar. 2021.
\newblock \href {http://arxiv.org/abs/2103.10730} {Muril: Multilingual representations for indian languages}.

\bibitem[{Kowsher et~al.(2022)Kowsher, Sami, Prottasha, Arefin, Dhar, and Koshiba}]{kowsher2022bangla}
Md~Kowsher, Abdullah~As Sami, Nusrat~Jahan Prottasha, Mohammad~Shamsul Arefin, Pranab~Kumar Dhar, and Takeshi Koshiba. 2022.
\newblock Bangla-bert: transformer-based efficient model for transfer learning and language understanding.
\newblock \emph{IEEE Access}, 10:91855--91870.

\bibitem[{Koyama et~al.(2021)Koyama, Hotate, Kaneko, and Komachi}]{koyama2021comparison}
Aomi Koyama, Kengo Hotate, Masahiro Kaneko, and Mamoru Komachi. 2021.
\newblock Comparison of grammatical error correction using back-translation models.
\newblock \emph{arXiv preprint arXiv:2104.07848}.

\bibitem[{Lan et~al.(2019)Lan, Chen, Goodman, Gimpel, Sharma, and Soricut}]{lan2019albert}
Zhenzhong Lan, Mingda Chen, Sebastian Goodman, Kevin Gimpel, Piyush Sharma, and Radu Soricut. 2019.
\newblock Albert: A lite bert for self-supervised learning of language representations.
\newblock \emph{arXiv preprint arXiv:1909.11942}.

\bibitem[{Lee et~al.(2017)Lee, Cho, and Hofmann}]{lee-etal-2017-fully}
Jason Lee, Kyunghyun Cho, and Thomas Hofmann. 2017.
\newblock \href {https://doi.org/10.1162/tacl_a_00067} {Fully character-level neural machine translation without explicit segmentation}.
\newblock \emph{Transactions of the Association for Computational Linguistics}, 5:365--378.

\bibitem[{Lin(2004)}]{lin-2004-rouge}
Chin-Yew Lin. 2004.
\newblock \href {https://aclanthology.org/W04-1013} {{ROUGE}: A package for automatic evaluation of summaries}.
\newblock In \emph{Text Summarization Branches Out}, pages 74--81, Barcelona, Spain. Association for Computational Linguistics.

\bibitem[{Liu et~al.(2019)Liu, Ott, Goyal, Du, Joshi, Chen, Levy, Lewis, Zettlemoyer, and Stoyanov}]{liu2019roberta}
Yinhan Liu, Myle Ott, Naman Goyal, Jingfei Du, Mandar Joshi, Danqi Chen, Omer Levy, Mike Lewis, Luke Zettlemoyer, and Veselin Stoyanov. 2019.
\newblock Roberta: A robustly optimized bert pretraining approach.
\newblock \emph{arXiv preprint arXiv:1907.11692}.

\bibitem[{M{\"a}ntyl{\"a} et~al.(2018)M{\"a}ntyl{\"a}, Graziotin, and Kuutila}]{mantyla2018evolution}
Mika~V M{\"a}ntyl{\"a}, Daniel Graziotin, and Miikka Kuutila. 2018.
\newblock The evolution of sentiment analysis—a review of research topics, venues, and top cited papers.
\newblock \emph{Computer Science Review}, 27:16--32.

\bibitem[{Ng et~al.(2014)Ng, Wu, Briscoe, Hadiwinoto, Susanto, and Bryant}]{ng-etal-2014-conll}
Hwee~Tou Ng, Siew~Mei Wu, Ted Briscoe, Christian Hadiwinoto, Raymond~Hendy Susanto, and Christopher Bryant. 2014.
\newblock \href {https://doi.org/10.3115/v1/W14-1701} {The {C}o{NLL}-2014 shared task on grammatical error correction}.
\newblock In \emph{Proceedings of the Eighteenth Conference on Computational Natural Language Learning: Shared Task}, pages 1--14, Baltimore, Maryland. Association for Computational Linguistics.

\bibitem[{Niu et~al.(2020)Niu, Yavuz, Zhou, Keskar, Wang, and Xiong}]{niu2020unsupervised}
Tong Niu, Semih Yavuz, Yingbo Zhou, Nitish~Shirish Keskar, Huan Wang, and Caiming Xiong. 2020.
\newblock Unsupervised paraphrasing with pretrained language models.
\newblock \emph{arXiv preprint arXiv:2010.12885}.

\bibitem[{Papineni et~al.(2002)Papineni, Roukos, Ward, and Zhu}]{papineni-etal-2002-bleu}
Kishore Papineni, Salim Roukos, Todd Ward, and Wei-Jing Zhu. 2002.
\newblock \href {https://doi.org/10.3115/1073083.1073135} {{B}leu: a method for automatic evaluation of machine translation}.
\newblock In \emph{Proceedings of the 40th Annual Meeting of the Association for Computational Linguistics}, pages 311--318, Philadelphia, Pennsylvania, USA. Association for Computational Linguistics.

\bibitem[{Powers(2020)}]{powers2020evaluation}
David~MW Powers. 2020.
\newblock Evaluation: from precision, recall and f-measure to roc, informedness, markedness and correlation.
\newblock \emph{arXiv preprint arXiv:2010.16061}.

\bibitem[{Price et~al.(2020)Price, Gifford-Moore, Fleming, Musker, Roichman, Sylvain, Thain, Dixon, and Sorensen}]{price2020six}
Ilan Price, Jordan Gifford-Moore, Jory Fleming, Saul Musker, Maayan Roichman, Guillaume Sylvain, Nithum Thain, Lucas Dixon, and Jeffrey Sorensen. 2020.
\newblock Six attributes of unhealthy conversation.
\newblock \emph{arXiv preprint arXiv:2010.07410}.

\bibitem[{Raffel et~al.(2020)Raffel, Shazeer, Roberts, Lee, Narang, Matena, Zhou, Li, and Liu}]{raffel2020exploring}
Colin Raffel, Noam Shazeer, Adam Roberts, Katherine Lee, Sharan Narang, Michael Matena, Yanqi Zhou, Wei Li, and Peter~J Liu. 2020.
\newblock Exploring the limits of transfer learning with a unified text-to-text transformer.
\newblock \emph{The Journal of Machine Learning Research}, 21(1):5485--5551.

\bibitem[{Reimers and Gurevych(2019)}]{reimers-2019-sentence-bert}
Nils Reimers and Iryna Gurevych. 2019.
\newblock \href {http://arxiv.org/abs/1908.10084} {Sentence-bert: Sentence embeddings using siamese bert-networks}.
\newblock In \emph{Proceedings of the 2019 Conference on Empirical Methods in Natural Language Processing}. Association for Computational Linguistics.

\bibitem[{Samia et~al.(2022)Samia, Rajee, Hasan, Faruq, and Paul}]{samia2022aspect}
Moythry~Manir Samia, Alimul Rajee, Md~Rakib Hasan, Mohammad~Omar Faruq, and Pintu~Chandra Paul. 2022.
\newblock Aspect-based sentiment analysis for bengali text using bidirectional encoder representations from transformers (bert).
\newblock \emph{International Journal of Advanced Computer Science and Applications}, 13(12).

\bibitem[{Sarker(2020)}]{Sagor_2020}
Sagor Sarker. 2020.
\newblock \href {https://github.com/sagorbrur/bangla-bert} {Banglabert: Bengali mask language model for bengali language understanding}.

\bibitem[{Sarker(2021)}]{sarker2021bnlp}
Sagor Sarker. 2021.
\newblock Bnlp: Natural language processing toolkit for bengali language.
\newblock \emph{arXiv preprint arXiv:2102.00405}.

\bibitem[{Shajalal and Aono(2018)}]{shajalal2018semantic}
Md~Shajalal and Masaki Aono. 2018.
\newblock Semantic textual similarity in bengali text.
\newblock In \emph{2018 International Conference on Bangla Speech and Language Processing (ICBSLP)}, pages 1--5. IEEE.

\bibitem[{Srivastava et~al.(2020)Srivastava, Makhija, and Gupta}]{srivastava2020noisy}
Ankit Srivastava, Piyush Makhija, and Anuj Gupta. 2020.
\newblock Noisy text data: Achilles’ heel of bert.
\newblock In \emph{Proceedings of the Sixth Workshop on Noisy User-generated Text (W-NUT 2020)}, pages 16--21.

\bibitem[{Sun and Jiang(2019)}]{sun2019contextual}
Yifu Sun and Haoming Jiang. 2019.
\newblock Contextual text denoising with masked language models.
\newblock \emph{arXiv preprint arXiv:1910.14080}.

\bibitem[{Wikipedia(2023)}]{wiki:benlan}
Wikipedia. 2023.
\newblock {List of languages by total number of speakers} --- {W}ikipedia{,} the free encyclopedia.
\newblock \url{https://en.wikipedia.org/wiki/List\_of\_languages\_by\_total\_number\_of\_speakers}.
\newblock [Online; accessed 13-June-2023].

\bibitem[{Xue et~al.(2020)Xue, Constant, Roberts, Kale, Al-Rfou, Siddhant, Barua, and Raffel}]{xue2020mt5}
Linting Xue, Noah Constant, Adam Roberts, Mihir Kale, Rami Al-Rfou, Aditya Siddhant, Aditya Barua, and Colin Raffel. 2020.
\newblock mt5: A massively multilingual pre-trained text-to-text transformer.
\newblock \emph{arXiv preprint arXiv:2010.11934}.

\bibitem[{Zaky and Romadhony(2019)}]{Zaky2019AnLS}
Damar Zaky and Ade Romadhony. 2019.
\newblock \href {https://api.semanticscholar.org/CorpusID:215800692} {An lstm-based spell checker for indonesian text}.
\newblock \emph{2019 International Conference of Advanced Informatics: Concepts, Theory and Applications (ICAICTA)}, pages 1--6.

\bibitem[{Zhang et~al.(2019)Zhang, Kishore, Wu, Weinberger, and Artzi}]{zhang2019bertscore}
Tianyi Zhang, Varsha Kishore, Felix Wu, Kilian~Q Weinberger, and Yoav Artzi. 2019.
\newblock Bertscore: Evaluating text generation with bert.
\newblock \emph{arXiv preprint arXiv:1904.09675}.

\end{thebibliography}
\bibliographystyle{acl_natbib}

\appendix
\section*{Appendix}
\section{Model Descriptions} \label{apA}
\subsection{Noise Identification} \label{model:nc}
\textbf{(a) SVM.} Support Vector Machine (SVM) is designed to find a hyperplane in a high-dimensional space. This hyperplane separates data points of different classes while maximizing the margin between these classes. For feature extraction, the TF-IDF Vectorizer was employed, utilizing both a character analyzer and a word analyzer. These are represented as SVM (C) for the character analyzer and SVM (W) for the word analyzer, respectively, using n-grams in the range of 1 to 4. Additionally, a combination of both character and word n-gram features was tested, denoted as SVM (C + W). \\
\textbf{(b) BiLSTM.} BiLSTM captures long-range dependencies and contextual information among items in a sequence. It has two LSTM layers, one that reads the input sequence in a forward direction and the other in a reverse direction. The outputs of these two layers are then concatenated to produce a final output for each item in the sequence. Our BiLSTM implementation features an embedding size of 512, a hidden size of 110, and consists of 2 layers.\\
\textbf{(c) Bangla-BERT-Base.} A pretrained Bangla language model using mask language modeling objective \citep{Sagor_2020}. It has the same architecture as the \verb|bert-base-uncased| \citep{DBLP:journals/corr/abs-1810-04805} model with an embedding size of 768 and a total parameter of 110M.

\subsection{Noise Reduction} \label{model:den}
\textbf{(a) BanglaT5.} A sequence-to-sequence transformer model that has been pre-trained using the span corruption objective \citep{bhattacharjee2022banglanlg}. It consists of 247 million parameters and has an embedding size of 768. For the implementation of the back-translation method, the BanglaT5 model, pre-trained on the BanglaNMT Bangla-English dataset \citep{bhattacharjee2022banglanlg}, is used for Bangla to English translation. Conversely, for English to Bangla translation, the BanglaT5 model pre-trained on the BanglaNMT English-Bangla dataset \citep{bhattacharjee2022banglanlg} is utilized. Additionally, the paraphrasing model employed by us is also BanglaT5 model, which has been pre-trained on the BanglaParaphrase dataset \citep{akil2022banglaparaphrase}.\\
\textbf{(b) BanglaBERT Generator.} This is an ELECTRA \citep{clark2020electra} generator that has been pre-trained using the Masked Language Modeling (MLM) objective, specifically on extensive Bangla corpora \citep{bhattacharjee-etal-2022-banglabert}. It has an embedding size of 768 and consists of 110M parameters. This model has been employed to perform the MLM task on out-of-vocabulary words and to execute random MLM with each word having a 20\% possibility of being masked. 

\subsection{Sentiment Analysis}
\textbf{(a) BanglaBERT.} An ELECTRA \citep{clark2020electra} discriminator model pre-trained with the Replaced Token Detection (RTD) objective. It has an embedding size of 768 and a total of 110M parameters \citep{bhattacharjee-etal-2022-banglabert}.\\
\textbf{(b) BanglaBERT Large.} A larger variant of BanglaBERT, with 335M parameters and an embedding size of 1024 \citep{bhattacharjee-etal-2022-banglabert}.\\
\textbf{(c) sahajBERT}\footnote{\url{https://huggingface.co/neuropark/sahajBERT}}\textbf{.} Pre-trained in Bangla language using Masked Language Modeling (MLM) and Sentence Order Prediction (SOP) objectives. It follows A Lite BERT (ALBERT) \citep{lan2019albert} architecture and has a total of 18M parameters and an embedding size of 128.\\
\textbf{(d) Bangla-Electra}\footnote{\url{https://huggingface.co/monsoon-nlp/bangla-electra}}\textbf{.} Trained with ELECTRA-small \citep{clark2020electra} with an embedding size of 128 and a total of 14M parameters.\\
\textbf{(e) MuRIL.} A BERT model pre-trained on 17 Indian languages and their transliterated counterparts \citep{khanuja2021muril}. It has 110M parameters and an embedding size of 768 for each token. The model is pre-trained on both monolingual and parallel segments.

\section{Performance Evaluation Metrics}\label{App:denoise}
\subsection{Noise Reduction} 
\textbf{(a) BLEU.} \textbf{B}i\textbf{L}ingual \textbf{E}valuation \textbf{U}nderstudy \citep{papineni-etal-2002-bleu} is a commonly used scoring method that measures the overlap between reference and candidate sentences, providing a similarity measurement.\\
\textbf{(b) ROUGE-L.} \textbf{R}ecall-\textbf{O}riented \textbf{U}nderstudy for \textbf{G}isting \textbf{E}valuation - \textbf{L}ongest Common Subsequence \citep{lin-2004-rouge} computes a similarity score by taking into account of longest common sub-sequences appearing in both reference and candidate sentences. Similar to the BLEU score, this scoring method does not provide much insight into semantic measurements, only the similarity of overlapping words/sub-sequences.\\
\textbf{(c) BERTScore.} BERTScore \citep{zhang2019bertscore} uses the cosine similarity of contextual embedding of the token provided from a BERT-based model. For this, we used the \verb|bert-score|\footnote{\url{https://pypi.org/project/bert-score/}} library, which uses a multilingual BERT for Bangla sentences.\\
\textbf{(d) SBERT Score.} For this method, we employed paraphrase-multilingual-MiniLM-L12-v2 \citep{reimers-2019-sentence-bert}, a model that maps sentences and paragraphs to a 384 dimensional dense vector space. It supports more than 50 languages and employs cosine similarity to assess the similarity between the input text and the ground truth.\\
\textbf{(e) BSTS.} \textbf{B}angla \textbf{S}emantic \textbf{T}extual \textbf{S}imilarity was first introduced by \citep{shajalal2018semantic}. It uses embeddings of Word2Vec to calculate the similarity between two sentences.\\
\textbf{(f) BERT-iBLEU.} The scoring method was originally proposed by \citep{niu2020unsupervised}, which combines BERT-Score and BLEU Score to measure the semantic similarity of sentences while penalizing for the presence of similar words. This scoring system is particularly suitable for our needs, as we intend to evaluate the method based on its ability to keep the semantic meaning intact while making necessary changes to reduce noises.\\
\textbf{(g) Word Coverage.} Pre-trained word embedding models like FastText \citep{sarker2021bnlp}, and Word2Vec \citep{sarker2021bnlp} create a vocabulary on the corpus they are trained on. As they are trained on noiseless sources like Wikipedia articles, their vocabulary contains accurate words. By measuring the percentage of tokens of our data covered in their vocabulary, we can gain insight into what percentage of tokens were noise reduced properly. However, this method may not address all types of noises. Additionally, we also calculated word coverage using the vocabulary of Bangla-BERT-Base \citep{Sagor_2020}.\\
\textbf{(h) Human Evaluation.} The output texts were evaluated by annotators by comparing them to the 1000 established ground truths. A noise reduced output was considered correct if it retained the same meaning as the ground truth and reduced at least some of the noise or complete noise from the original sentence. In essence, the score represents the proportion of accurate noise reduced data relative to the 1000 ground truth. The score can be defined as:
\[
\text{Score (Human Evaluation)} = \frac{x}{T} * 100
\]
Here, x = Accurately  noise reduced  data\\
T = Total  number  of  data

\subsection{Classification} \label{App:class}
For both classification tasks (noise and sentiment), we used micro precision, recall, and F1-score.\\
\textbf{(a) Precision.} Precision measures the accuracy of positive predictions, specifically how many of them are correct (true positives) \citep{powers2020evaluation}. Alternatively known as True Positive Accuracy (TPA), it is calculated as:
\[
\text{Precision} = \frac{TP}{TP + FP}
\]
 where TP indicates true positive and FP indicates false positive.\\
\textbf{(b) Recall.} Recall, or True Positive Rate (TPR), gauges the classifier's ability to accurately predict positive cases by determining how many of them it correctly identified out of all the positive cases in the dataset \citep{powers2020evaluation}. It is defined as:
\[
\text{Recall} = \frac{TP}{TP + FN}
\]
 where TP indicates true positive and FN indicates false negative.\\
\textbf{(c) F1-Score.} The F1-score is the harmonic mean of precision and recall, providing a balance between the two in cases where one may be more significant than the other.
F1-score is defined as:
\[
\text{F1-Score} = 2 \times \frac{\text{Precision} \times \text{Recall}}{\text{Precision} + \text{Recall}}
\]

\onecolumn
\section{Types of Noise in NC-SentNoB}\label{ap1}
NC-SentNoB dataset contains labeled data for 10 types of noise. Table \ref{tab:noise_example} illustrates the definition of each noise type annotators used for the annotation process. In case of \textbf{Punctuation Error}, an exception was made for sentences that end without a period "{\bng .}" due to the nature of the data. If such instances were considered errors, the majority of the data would be labeled as having punctuation errors. This could lead to trained models predominantly focusing on this single type of error, rather than recognizing and learning from a broader range of punctuation errors.

\begin{table}[h]
    \centering
   
    \resizebox{\textwidth}{8.5cm}{
    \begin{tabular}{p{0.2\textwidth}p{0.4\textwidth}p{0.4\textwidth}}
         \hline
         \multicolumn{1}{c}{\textbf{Type}} & \multicolumn{1}{c}{\textbf{Definition}} & \multicolumn{1}{c}{\textbf{Example with Correction}} 
         \\ \hline
         \textbf{Local Word}
         & Any regional words even if there is a spelling error 
         & \textbf{{[}N{]}} {\bng pResh/nr saeth Ut/terr ekan iml \textcolor{red}{pa{I}lam} na}\newline\textbf{{[}C{]}} {\bng pResh/nr saeth Ut/terr ekan iml \textcolor{ForestGreen}{eplam} na}\newline\textbf{{[}E{]}} I did not find any similarity between the question and the answer.
         \\ \hline
         \textbf{Word Misuse} 
         & Wrong use of words or unnecessary repetitions of words 
         & \textbf{{[}N{]}} {\bng taek Aa{I}enr Aa{O}tay \textcolor{red}{shain/t} ed{O}ya eHak}\newline\textbf{{[}C{]}} {\bng taek Aa{I}enr Aa{O}tay \textcolor{ForestGreen}{shais/t} ed{O}ya eHak}\newline\textbf{{[}E{]}} He should be punished under the law.\\ \hline
         \textbf{Context/Word missing} 
         & Not enough information or missing words 
         &\textbf{{[}N{]}} {\bng itin EkmaE paern E{I} mHaibpd \fcolorbox{red}{white}{\textcolor{white}{-}} prRithbiiek rkKa kret}\newline\textbf{{[}C{]}} {\bng itin EkmatR paern E{I} mHaibpd \textcolor{ForestGreen}{ethek} prRithbiiek rkKa kret}\newline\textbf{{[}E{]}} He is the only one who can save the world from this catastrophe.
         \\ \hline
         \textbf{Wrong Serial} 
         & Wrong order of the words 
         & \textbf{{[}N{]}} {\bng saraedesh Apradhii khNNujun , Aaera \textcolor{red}{Hey HenY}} \newline\textbf{{[}C{]}} {\bng Aaera \textcolor{ForestGreen}{HenY Hey} saraedesh Apradhii khNNujun} \newline\textbf{{[}E{]}} Search for the criminal desperately.
         \\ \hline
         \textbf{Mixed Language} 
         & Words in another language. Foreign words that were adopted into the Bangla language over time are excluded from this type. 
         &\textbf{{[}N{]}} {\bng bha{I}er E{I} \textcolor{red}{inUjTa} esra \textcolor{red}{in{I}j}}\newline\textbf{{[}C{]}} {\bng bha{I}er, E{I} \textcolor{ForestGreen}{khbrTa} esra \textcolor{ForestGreen}{khbr}}\newline\textbf{{[}E{]}} Brother, this news is the best news.
         \\ \hline
         \textbf{Punctuation Error} & Improper placement or missing punctuation. Sentences ending without "{\bng .}" ({\bng dNNairh}) were excluded from this type.  & \textbf{{[}N{]}} {\bng perr par/Tguela keb Aaseb bha{I} \fcolorbox{red}{white}{\textcolor{white}{1}}}\newline\textbf{{[}C{]}} {\bng perr pr/bguela keb Aaseb bha{I}\textcolor{ForestGreen}{?}}\newline\textbf{{[}E{]}} When will the next episodes air brother?
         \\ \hline
         \textbf{Spacing Error} & Improper use of white space & \textbf{{[}N{]}} {\bng \textcolor{red}{prhaeshana Ta} cailey egel bhaela Heta}\newline\textbf{{[}C{]}} {\bng \textcolor{ForestGreen}{prhaeshanaTa} cailey egel bhaela Heta}\newline\textbf{{[}E{]}} It would be better to continue studying
         \\ \hline
         \textbf{Spelling Error} 
         & Words not following spelling of Bangla Academy Dictionary 
         & \textbf{{[}N{]}} {\bng \textcolor{red}{baibek} Et \textcolor{red}{jal} kha{O}yaena iThk na}\newline\textbf{{[}C{]}} {\bng \textcolor{ForestGreen}{bhabiiek} Et \textcolor{ForestGreen}{jhal} kha{O}yaena iThk na}\newline\textbf{{[}E{]}} It is not right to feed the sister-in-law so much spice.
         \\ \hline
         \textbf{Coined Word} 
         & Emoji, symbolic emoji, link 
         & \textbf{{[}N{]}} {\bng Aaeg janel Aapnar saeth edkha krtam} \fcolorbox{red}{white}{\scalerel*{\includegraphics{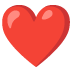}}{X}} \newline 
         \textbf{[C]} \textcolor{red}{\ding{55}} \newline
         \textbf{{[}E{]}} If I knew I would've met you earlier \scalerel*{\includegraphics{heart.png}}{X} \\ \hline
         \textbf{Others} 
         & Noises that do not fall into categories mentioned above. 
         & \textbf{{[}N{]}} {\bng red kut/tar bac/caedr phNNais ca{I}}\newline
         \textbf{[C]} \textcolor{red}{\ding{55}} \newline
         \textbf{{[}E{]}} I want those sons of bitches hanged. 
         \\ \hline
    \end{tabular}
    }
    \caption{Types of noise with the definition that was used to annotate the dataset. \textbf{N} represents the original noisy sentence, \textbf{C} represents the corrected sentence, and \textbf{E} represents the corresponding English translation. The types \textbf{Coined Word}, and \textbf{Others} do not have any correction as these types of noise are essential to the meaning of the sentence. For each example, noisy words of that particular type are marked with \textcolor{red}{Red} color, and their correction is marked with \textcolor{ForestGreen}{Green} color.}
    \label{tab:noise_example}
\end{table}

\clearpage

\section{Failure Cases of Back-translation}\label{ap2}
To provide insight into the performance drop, we have illustrated examples where the back-translation method using Google Translate fails to adequately reduce noise in the input text in table \ref{tab:bt_scen}. Moreover, it often alters or completely removes important contextual words, which possibly impacts the performance of sentiment analysis. Given a human evaluation score of 37.90\%, it can be said that back-translation via Google Translate fails to effectively correct more than 50\% of the 1000 manually corrected data.

\begin{table}[h]
    \centering
    \resizebox{\textwidth}{8.74cm}{
    \begin{tabular}{p{0.5\textwidth} p{0.5\textwidth}}
        \hline
        \multicolumn{1}{c}{\textbf{Noisy data and corresponding Back-Translation}} & 
        \multicolumn{1}{c}{\textbf{Observation}}
        \\ \hline
        \textbf{[N]} {\bng E{I} juyar Taka papn \textcolor{red}{ept} Aamar men Hy} \newline
        \textbf{[C]} {\bng E{I} juyar Taka papn epeta Aamar men Hy} \newline
        \textbf{[E]} I think the gambling money went to Papan. \newline
        \textbf{[B]} {\bng Aaim men kir E{I} juyar Taka \textcolor{red}{pireshadh kra Heb}} \newline
        \textbf{[BE]} I think this gambling money will be repaid.
        & 
        The input text contained only a spelling mistake, but the back-translation introduced new words, removed a named entity, and altered the sentence's meaning.
        \\ \hline

        \textbf{[N]} {\bng bha{I} khabaerr \textcolor{red}{sadh} ejmin eHak na ekn Aapnar muekh egel esTa \textcolor{red}{AimRt} Hey Jay \textcolor{red}{dhn/nbad}} \newline 
        \textbf{[C]} {\bng bha{I} khabaerr sWad eJmin eHak na ekn Aapnar muekh egel esTa AmrRt Hey Jay, dhnYbad} \newline
        \textbf{[E]} Brother, whatever the taste of the food is, it becomes necter in your mouth, thanks.\newline
        \textbf{[B]} {\bng bha{I}, khabaerr \textcolor{ForestGreen}{sWad} Ja{I} eHak na ekn, \textcolor{red}{ETa Aapnar muekh Aaech}.} \newline
        \textbf{[BE]} Brother, whatever the taste of the food is, it's in your mouth.
        & 
        The input text had multiple spelling mistakes and punctuation errors. The back-translation corrected one of these errors but changed the meaning of part of the sentence.
        \\ \hline
        
        \textbf{[N]} {\bng Eedr \textcolor{red}{ipeTr} camrha etala Heb} \newline 
        \textbf{[C]} {\bng Eedr ipeThr camrha etala Heb} \newline
        \textbf{[E]} Their backs will be skinned.\newline
        \textbf{[B]} {\bng tara camrha Heb} \newline
        \textbf{[BE]} They will become skin.
        & 
        The input text contained only a spelling mistake. However, the back-translation removed contextually important words, rendering the sentence meaningless.
        \\ \hline
        
        \textbf{[N]} {\bng \textcolor{red}{A amar emechr} saeth E{I} eHaeTl} \newline 
        \textbf{[C]} {\bng Aamar emesr saeth E{I} eHaeTl} \newline
        \textbf{[E]} This hotel is with my hostel.\newline
        \textbf{[B]} {\bng Aamar \textcolor{red}{jal} idey E{I} eHaeTl} \newline
        \textbf{[BE]} This hotel with my net.
        & 
        The back-translation altered a keyword in the sentence, which resulted in a loss of meaning.
        \\ \hline
        
        \textbf{[N]} {\bng \textcolor{red}{ilbur saet} ik Aada ekhet Heb }\fcolorbox{red}{white}{\textcolor{white}{1}} \newline
        \textbf{[C]} {\bng elbur saeth ik Aada ekhet Heb?} \newline
        \textbf{[E]} Do I need to eat ginger with lemon?\newline
        \textbf{[B]} {\bng \textcolor{red}{Aaim} ik} \textcolor{red}{Libur Sate} {\bng E Aada kha{O}ya Uict\textcolor{ForestGreen}{?}} \newline
        \textbf{[BE]} Should I eat ginger with Libur Sate?
        & 
        The back-translation failed to correct a spelling mistake and converted the word into English, but it successfully added the missing punctuation.
        \\ \hline
        
        \textbf{[N]} {\bng rana edr mt echelra jaet Hairey na jay} \newline 
        \textbf{[C]} {\bng ranaedr mt echelra Jaet Hairey na Jay} \newline
        \textbf{[E]} So that boys like Rana don't get lost. \newline
        \textbf{[B]} {\bng ranar meta echelra \textcolor{red}{eres} eHer Jay na} \newline
        \textbf{[BE]} Boys like Rana do not lose in race.
        & 
        The input sentence had spacing and spelling errors. The back-translation fixed the spacing issue but introduced mixed language, changing the sentence's meaning.

        \\ \hline
    \end{tabular}
    }
    \caption{Example scenarios where back-translation with google translate fails to reduce noise in the text. \textbf{N} represents the original noisy sentence, \textbf{C} represents the corrected sentence, \textbf{E} represents its English translation, \textbf{B} represents the result of back-translation, and \textbf{BE} represents the direct English translation of back-translated output. For each example, noisy words are marked with \textcolor{red}{Red} color and noise reduced words are marked with \textcolor{ForestGreen}{Green} color.
    }
    \label{tab:bt_scen}
\end{table}
\end{document}